\documentclass[12pt]{article}
\usepackage{graphicx}
\usepackage{natbib} %comment out if you do not have the package
\usepackage{url} % not crucial - just used below for the URL 

\newcommand{\blind}{0}

\usepackage{bm}
\usepackage{mathtools} % vertical centered colon (vcentcolon)
\usepackage{amsmath} % For \argmax
\DeclareMathOperator*{\argmax}{argmax} % thin space, limits underneath in displays
\usepackage{amssymb}
\usepackage{lineno}

\usepackage{subdepth} % subscript would be at the same height no matter whether there is a superscript.
\usepackage{graphicx}
\usepackage{caption} % use for subfigure
\usepackage{subcaption} % use for subfigure
\usepackage[toc,page]{appendix}
\usepackage{float}

\usepackage{bookmark}
\bookmarksetup{open,numbered}

\usepackage{tabularx}
\newcolumntype{Y}{>{\centering\arraybackslash}X}
\usepackage{multirow}

\usepackage{booktabs} % For thick horizontal rule
\usepackage[shortlabels]{enumitem} % For enumerate labels
\usepackage{changes} % Tracking changes in different version in latex, see page: https://tex.stackexchange.com/questions/65453/track-changes-in-latex
\usepackage{algorithm}
\usepackage{algpseudocode}
\usepackage{setspace} % Change the space inside an algorithm

\definechangesauthor[name={Per cusse}, color=brown]{Kungang}
\definechangesauthor[name={Per cusse}, color=orange]{Anh}
\definechangesauthor[name={Per cusse}, color=red]{Apley}

\begin{document}

\def\spacingset#1{\renewcommand{\baselinestretch}%
{#1}\small\normalsize} \spacingset{1}

%%%%%%%%%%%%%%%%%%%%%%%%%%%%%%%%%%%%%%%%%%%%%%%%%%%%%%%%%%%%%%%%%%%%%%%%%%%%%%

\if0\blind
{
  \title{\bf Concept Drift Monitoring and Diagnostics of Supervised Learning Models via Score Vectors}
  \author{Kungang Zhang\textsuperscript{1} \\
%  \thanks{The authors gratefully acknowledge \textit{please remember to list all relevant funding sources in the unblinded version}}\hspace{.2cm}\\
    Anh T.\ Bui\textsuperscript{2} \\
    Daniel W.\ Apley\textsuperscript{1} \\
    \textsuperscript{1}Department of Industrial Engineering and Management Sciences,\\ Northwestern University \\
    \textsuperscript{2}Department of Statistical Sciences and Operations Research,\\ Virginia Commonwealth University
    }
  \maketitle
} \fi

\if1\blind
{
  \bigskip
  \bigskip
  \bigskip
  \begin{center}
    {\LARGE\bf Concept Drift Monitoring and Diagnostics of Supervised Learning Models via Score Vectors}
\end{center}
  \medskip
} \fi

\bigskip
\begin{abstract}
Supervised learning models are one of the most fundamental classes of models. Viewing supervised learning from a probabilistic perspective, the set of training data to which the model is fitted is usually assumed to follow a stationary distribution. However, this stationarity assumption is often violated in a phenomenon called concept drift, which refers to changes over time in the predictive relationship between covariates $\bm{X}$ and a response variable $Y$ and can render trained models suboptimal or obsolete. We develop a comprehensive and computationally efficient framework for detecting, monitoring, and diagnosing concept drift. Specifically, we monitor the Fisher score vector, defined as the gradient of the log-likelihood for the fitted model, using a form of multivariate exponentially weighted moving average, which monitors for general changes in the mean of a random vector. In spite of the substantial performance advantages that we demonstrate over popular error-based methods, a score-based approach has not been previously considered for concept drift monitoring. Advantages of the proposed score-based framework include applicability to {broad classes of parametric models}, more powerful detection of changes as shown in theory and experiments, and inherent diagnostic capabilities for helping to identify the nature of the changes.
\end{abstract}

\noindent%
\vfill

\newpage
\spacingset{2} % DON'T change the spacing!
%\spacingset{1} % DON'T change the spacing!
\section{Introduction}
\label{s_CD:intro}
In supervised learning, models are trained to predict a response variable $Y$, given an observed set of covariates $\bm{X} \in \mathbb{R}^p$. The desire is usually to make test prediction accuracy metrics~(i.e., $R^2$, classification accuracy, F1-score, etc.) as high as possible. From a probabilistic perspective, a training sample of observations, $\{\bm {x}_i, y_i\}_{i=1}^n$, is usually assumed to be drawn from some joint distribution, $P (\bm {X}, Y)$, such that the conditional distribution, $P(Y|\bm{X};\bm{\theta})$, is stationary across the $n$ observations~\citep{hulten2001mining}. Here, {$\bm {x}_i \in \mathbb{R}^{p}$ and $y_i \in \mathbb{R}$ are the covariates and response variable in the $i^{\mathrm{th}}$ observation respectively} {(The case with scalar response can easily be extended to a vector response)}, and we have parametrized the conditional distribution via the vector $\bm{\theta}$. However, the stationarity assumption is often violated in real applications, a phenomenon known as concept drift~\citep{moreno2012unifying,vzliobaite2016overview}. For example, models used to predict customers' probabilities of defaulting by credit-scoring agencies are usually fitted to training data collected over a three to five years period, so that changing financial environments may result in a partially-outdated predictive relationship by the time the trained model is used to score new customers~\citep{crook1992degradation,vzliobaite2016overview}. In this work, we follow what appears to be the most common terminology according to~\citep{moreno2012unifying} and~\citep{vzliobaite2016overview}. In general, any temporal drift in the joint distribution $P(\bm {X}, Y)$ is called ``data set drift". Decomposing the joint distribution as $P(\bm{X}, Y) = P(Y|\bm{X})P(\bm {X})$, the term ``concept drift" refers to temporal drift in $P(Y|\bm{X})$~(which is what the fitted supervised learning model approximates), while ``covariate drift" refers to temporal drift in $P(\bm{X})$. Our approach is intended to detect drift in the predictive relationship $P(Y|\bm {X})$ and not in $P(\bm{X})$ (see Section~\ref{s_CD:decou_cd} for elaboration). For detecting changes in $P(\bm{X})$ or in $P(\bm{X})$ and $P(Y|\bm {X})$ together, we refer readers to~\citep{raza2015ewma} and \citep{mejri2018new, mejri2021new}.

Concept drift poses many challenges in constructing trustworthy supervised learning models. Concept drift during both training~(for historical data) and model usage~(for future data) can degrade performance of supervised learning models: during training, concept drift results in there being no single predictive model $P(Y|\bm {X}; \bm{\theta})$, since the model is changing over time; while concept drift during usage degrades the accuracy of the predictive model, relative to what it could be with an updated model for $P(Y|\bm{X};\bm{\theta})$. In order to obtain and maintain the highest possible predictive performance of supervised learning models, \textit{retrospective} concept drift analysis of the training data and \textit{prospective} concept drift monitoring of the future data to which the model is applied are important and should be a standard component of the predictive modeling process.

Concept drift is common in practice and often originates from changes of some hidden effects in generating the response outcomes of interest~\citep{tsymbal2008dynamic,vzliobaite2012beating,widmer1996learning,kukar2003drifting,donoho2004early,carmona2010gnusmail,harel2014concept,fong2015change,vzliobaite2016overview}. For example, in the problem of antibiotic resistance in nosocomial infections, supervised learning models are trained and validated using a data set, with response labels indicating whether the level of sensitivity of a pathogen to an antibiotic is sensitive, resistant, or intermediate, and the covariates being patients' demographical data and conditions of hospitalization~\citep{pechenizkiy2005knowledge}. After properly training and validating such a model, it can with high accuracy determine whether or not a pathogen is sensitive to an antibiotic or not. However, according to medical experts~\citep{kukar2003drifting}, hidden effects due to failures and/or replacements of some medical equipment, changes in personnel, and seasonal bacterial outbreaks could result in unexpected changes over time in resistance of new pathogen strains to antibiotics. In other words, there is likely a significant level of concept drift in nosocomial infections. Consequently, bacteria that are predicted to be sensitive to a particular antibiotic may now be resistant, and the errors of the predictive model become unacceptable. 

The problem of concept drift is gaining increasing attention, because increasingly data are organized in the form of data streams, and it is often unrealistic to expect that data distributions remain stable for a long period of time. Most state-of-the-art concept drift detection and adaptation methods, like ensemble methods~\citep{wang2003mining}, neural networks~\citep{calandra2012learning}, and other non-parametric methods~\citep{bifet2007learning,frias2015online}, are based on monitoring the classification error or some metrics derived from it~\citep{ross2012exponentially,gonccalves2014comparative,barros2018large}. 

The main purpose of this paper is to introduce a new and comprehensive concept drift framework that is based on monitoring the Fisher scores of the individual observations over time, as opposed to metrics derived from their {prediction} errors. {Existing concept drift methods that are based on monitoring the error rate may fail to detect concept drift, because a change in $P(Y|\bm{X};\bm{\theta})$ does not necessarily result in an increase~(or a decrease) in the error rate~(see Section~\ref{ss_CD:score_func}). Note that a change in $P(Y|\bm{X};\bm{\theta})$ that does not change the error rate still means that, if the model is updated accordingly after concept drift detection, the predictive accuracy of the model can be improved~({see} supplementary Section~B.1). Thus it is still desirable to detect such changes.} The Fisher score, which we will refer to as simply the score, of an observation $(\bm{X}, Y)$ is defined as the gradient of the log-likelihood, $\log{P(Y|\bm{X};\bm{\theta})}$, with respect to the parameters, $\bm{\theta}$. As shown in the following sections, comparisons between the score-based and error-based methods from both theoretical and empirical perspectives provide strong justification for a score-based approach. The score-based framework that we develop and demonstrate is intended to detect, monitor, and diagnose concept drift with the following major advantages and attractive properties.

\begin{itemize}
\item
\textit{The approach is based on monitoring for changes in the mean of the score vector, which has strong theoretical justification since theory dictates that the mean of the score vector changes if and only if $P(Y|\bm{X};\bm{\theta})$ changes, under fairly general conditions~(Section~\ref{ss_CD:score_func}).}
\item
{\textit{The score-based approach can detect concept drift even when the error rate doesn't change. Upon the detection of the concept drift, the predictive model can be updated giving a lower prediction error rate.}}
\item
\textit{The score-based approach is more sensitive to changes in $P(Y|\bm{X}; \bm{\theta})$ and thus more quickly detects concept {drift} than error-based methods, as demonstrated empirically with a number of examples~(Section~\ref{s_CD:real_data} and {supplementary Section~B.2 and~B.4}).}
\item
\textit{The score-based approach is applicable to {broad classes of} parametric classification or regression models that can be interpreted probabilistically in terms of $P(Y|\bm{X};\bm{\theta})$, parameterized by a vector of parameters $\bm{\theta}$.}
\item
\textit{The score-based approach also naturally provides a convenient means of diagnosing the nature of the concept drift~(e.g., which parameters have changed) as derived in Section~\ref{s_CD:decou_cd} and demonstrated in Section~\ref{s_CD:real_data} and {supplementary Sections~B.4 and~C}.}
%\item
%\textit{The score-based perspective converts the concept drift problem to the problem of monitoring changes in the mean of a random vector, for which many existing multivariate statistical process control~(SPC) methods are available. We focus on a multivariate exponentially weighted moving average~(MEWMA) control chart~(\citep{lowry1992multivariate,hotelling1947multivariate,montgomery2007introduction}), which has desirable characteristics for our problem~(see Section~\ref{ss_CD:MEWMA}), but as new methods are developed they can also be applied.}
%\item
%\textit{The computations involved in the score-based approach are almost the same computations involved in stochastic gradient descent~(SGD) algorithms, which are increasingly commonly used to fit parametric supervised learning models. In this sense, the computations come at very little additional cost, resulting in a computationally inexpensive approach~(Section~\ref{ss_CD:sgd_score}).}
\item
\textit{The same score-based concepts can be used in a consistent manner either retrospectively~(e.g., after fitting a supervised learning model to a set of training data, to validate that the training data were stable and, if not, provide diagnostic information as to why) or prospectively~(e.g., when using a fitted supervised learning model to predict new cases, to quickly signal when the predictive model has changed, indicating that it is time to update the model).}
\end{itemize}

We elaborate on these properties in Sections~\ref{s_CD:theory_analysis_score} and~\ref{s_CD:decou_cd} and provide {an illustrative example with real data (credit risk scoring during the subprime mortgage crisis) in Section 5. The Supplementary Materials sections provide additional real and simulation comparison examples and derivations of various results.} 

%simulation results demonstrating them in \replaced[id=Kungang, remark={}]{supplementary Sections~C and~D}{Appendices~\ref{app:simu_MRL} and~\ref{app:cd_diag}}. Two real examples \replaced[id=Kungang, remark={}]{(credit risk scoring in Section~\ref{s_CD:real_data} and bike sharing demand prediction in supplementary Section~B)}{(credit risk scoring and bike sharing demand prediction) in Section~\ref{s_CD:real_data}} serve as case-studies to further illustrate its usage.

\section{Relation to Prior Work}
\label{s_CD:review}
In this section, we briefly review relevant existing literature on monitoring different types of drift in data distributions, in terms of goals and methodologies. 
%{{In the literature on data set drift, the terminology is not always consistent. In this work, we follow what appears to be the most common terminology according to~\citep{moreno2012unifying} and~\citep{vzliobaite2016overview}. In general, any temporal drift in the joint distribution $P(\bm {X}, Y)$ is called ``data set drift". Decomposing the joint distribution as $P(\bm{X}, Y) = P(Y|\bm{X})P(\bm {X})$, the term ``concept drift" refers to temporal drift in $P(Y|\bm{X})$~(which is what the fitted supervised learning model approximates), while ``covariate drift" refers to temporal drift in $P(\bm{X})$.}}
In general, the methods in the concept drift literature can be categorized into two classes~\citep{tsymbal2004problem,harel2014concept}: 1) model adaptation~(or online learning) methods and 2) concept drift detection methods. Model adaptation methods mainly focus on maintaining the performance of machine learning models in the presence of concept drift, without formally detecting or diagnosing the drift~\citep{wang2003mining,tsymbal2008dynamic,gonccalves2014comparative,barros2018large}. To maintain a good prediction metric related to classification or regression error, models are automatically updated~(i.e., adapted) online continuously as new observations are collected, which is sometimes called online or incremental learning. This class of methods is not particularly relevant to our work, since our goal is concept drift detection and diagnosis, and not model adaptation. In fact, we view our approach as something that can be used in conjunction with model adaptation methods to make them more efficient and interpretable. In particular, the model adaptation could be turned on only when the concept drift detection component has indicated that $P(Y|\bm{X};\bm{\theta})$ has changed.

%\deleted[id=Kungang, remark={}]{, and examples in this category include DDM~(\citep{gama2004learning}), EDDM~(\citep{baena2006early}), ECDD~(\citep{ross2012exponentially}), and Linear-4-rate~(\citep{wang2015concept}), etc}

The concept drift detection methods are more relevant to our work. Although most of these methods were originally proposed in a continuous model adaptation setting, we focus on their use and performance in our setting, in which one has a single fixed model fitted to one set of training data and applies it to usage data without updating it, so that small and gradual concept drift can be more easily detected. The drift detection method~(DDM)~\citep{gama2004learning} monitors the accumulated classification error rate as it evolves over time. The early drift detection method~(EDDM)~\citep{baena2006early} instead monitors the intervals between consecutive errors. The {exponentially weighted moving average (EWMA)} for concept drift detection~(ECDD)~\citep{ross2012exponentially} method monitors the misclassification rate using a univariate EWMA control chart. In the Linear-4-rate method~\citep{wang2015concept}, four statistics are monitored simultaneously to detect concept drift in binary classification problems. Most of the methods are designed specifically for binary classification and based on simple error-based quantities, like classification error, precision, recall, and residuals of the classification model~\citep{wang2015concept,barros2018large}. As we demonstrate throughout the paper, our score-based approach is far more powerful than error-based approaches.

Although the concept drift community seems to be unaware of the potential of score-based approaches for concept drift~(e.g., it is not mentioned in the recent surveys in~\citep{barros2018large} and~\citep{lu2018learning}), there have been some works in the econometrics literature that have used the score vector to test for changes in the parameters of regression models~\citep{kuan1995generalized,zeileis2005unified,zeileis2007generalized,xia2009monitoring}. Our work differs from this prior work in that we develop a comprehensive framework for{,} and investigate issues more relevant in{,} the typical situations to which the so-called concept drift paradigm refers. The econometrics work is developed mainly around the change-point paradigm in which it is assumed there is a single point in time at which $\bm{\theta}$ changes from some before-value to some after-value, and the goal is to identify the change point with formal hypotheses testing. In contrast, our approach is developed around the much more general and flexible situation in which $\bm{\theta}$ can continuously drift and/or change abruptly at multiple change-points, which is far more common in typical concept drift applications. Ours is more of an exploratory approach to inform the predictive modeling process, as opposed to formal {hypothesis} testing of well-defined but restrictive hypotheses. Our approach is also more general in the sense that it applies to broad classes of parametric supervised learning models in which $\log{P(Y|\bm{X};\bm{\theta})}$ is differentiable in $\bm{\theta}$~(e.g., generalized linear models, neural networks, Gaussian process models, etc.). Moreover, we develop and study a number of other aspects that are highly relevant to the concept drift setting, including a diagnostic framework for understanding the nature of the concept drift.

\section{Monitoring the Score Vector for Concept Drift}
\label{s_CD:theory_analysis_score}
To monitor supervised learning models for concept drifts, we propose a systematic framework based on the sample score vectors derived from parametric models. {In Section~\ref{ss_CD:score_func}}, we present theoretical arguments for using the score vector as the basis for concept drift monitoring. {In Section~\ref{ss_CD:sgd_score}}, we explain empirical counterparts to the theoretical arguments, including interpretations when the parametric model structure is only an approximation to the true $P(Y|\bm{X})$, and connections to {stochastic gradient descent (SGD)} algorithms. {In Section~\ref{ss_CD:MEWMA}, we develop} the {multivariate EWMA} (MEWMA) procedure for monitoring the mean of the score vector, {and in Section~\ref{ss_CD:high_dim_score}} we discuss various implementation issues and how to handle high-dimensional $\bm{\theta}$ and regularized models. Algorithm~\ref{alg_CD:score_based_algo} provides high-level pseudocode for the entire procedure, details of which are explained in the subsequent sections.

\begin{algorithm}
\setstretch{1}
\caption{Score-Based Concept Drift Monitoring and Diagnostics}\label{alg:cd_score_based}
\begin{algorithmic}[1]
\State \textbf{Step 1: Retrospective Analysis} 
\State Set $\bm{Stable}=\bm{False}$
\While {$\bm{Stable}=\bm{False}$}
\State Fit a supervised learning model to training data $\{(\bm{x}_i, y_i)\}_{i=1}^n$.
\State Calculate score vectors (Eq.~\ref{eqn_CD:score_func}) for $\{(\bm{x}_i, y_i)\}_{i=1}^n$ and apply the MEWMA (Sec.~\ref{ss_CD:MEWMA}).
\If {the MEWMA indicates the training data are stable}
\State Set $\bm{Stable}=\bm{True}$
\Else
\State Discard the portion of training data over which the drift occurred and denote the remaining subset of training data as $\{(\bm{x}_i, y_i)\}_{i=1}^{n}$ with $n$ reduced.
\EndIf
\EndWhile

\State \textbf{Step 2: Concept Drift Monitoring}
\State Partition the stable data set from Step 1, $\{(\bm{x}_i, y_i)\}_{i=1}^n$, into $\mathcal{D}_1\vcentcolon=\{\bm {x}_i, y_i\} _{i=1} ^{n_1}$ and $\mathcal{D}_2\vcentcolon=\{\bm {x}_i, y_i\} _{i=n_1+1} ^{n}$.
\State Fit a new parametric supervised learning model $\mathcal{M}$ to $\mathcal{D}_1$.
\State Calculate the score vectors (Eq.~\ref{eqn_CD:score_func}) for $\mathcal{D}_2$ using $\mathcal{M}$.
\State Compute MEWMA statistics $\{\bm {z}_t\}_{t=n_1+1}^n$ and $\{T_t^2\}_{t=n_1+1}^n$ (Eq.~\ref{eqn_CD:hotellingt2}) for $\mathcal{D}_2$ using $\mathcal{M}$.
\State For a desired false alarm rate $\alpha$, set $UCL$ as the $1-\alpha$ sample quantile of $\{T_t^2\}_{t=n_1+1}^n$.
\State For Phase-II ``on-line" data $\tilde{\mathcal{D}}\vcentcolon=\{\tilde{\bm{x}}_i,\tilde{y}_i\}_{i = 1}^{\tilde{n}}$, calculate the score vectors and apply the MEWMA with the $UCL$ from Line $17$ to monitor and detect concept drift.

\State \textbf{Step 3: Concept Drift Diagnosis After Concept Drift Detection}
\State Plot the $T_t^2$ statistics over time for the data over which concept drift was detected.
\State Decouple the concept drift in the score vectors by calculating the Fisher-decoupled score vectors $\tilde{\bm{s}}_t \vcentcolon= \widehat{\mathbb {I}}^{-1}(\bm { \theta}^{ (0)})\bm{s}_t$  (see line following Eq.~\ref{eqn_CD:uniewma-decoupled}).
\State Plot the univariate EWMA charts for each component of Fisher-decoupled score vectors $\tilde{\bm{s}}_t$.

\end{algorithmic}
\label{alg_CD:score_based_algo}
\end{algorithm}

\subsection{The Score Vector as a Basis for Concept Drift Monitoring}
\label{ss_CD:score_func}
Supervised learning is used to approximate an underlying conditional response distribution, which, if parametric, can be denoted as $P(Y|\bm{X};\bm{\theta})$. For example, for a regression neural network, the conditional mean $E[Y|\bm{X};\bm{\theta}]$ is modeled as a neural network, and $Y$ is assumed to be its conditional mean plus~(typically) a Gaussian error; or for a classification neural network, $Y$ is multinomial with class probabilities that are modeled as a neural network. Fitting the model then entails estimating the parameters $\bm{\theta}$ by maximizing the log-likelihood, which can be viewed as an empirical approximation to $E_{\bm{\theta}^{(0)}}[\log{P(Y|\bm{X};\bm{\theta})}]$, where $\bm{\theta}^{(0)}$ denotes the true parameters and the subscript $\bm { \theta} ^{ (0)}$ on the expectation operator indicates that it is {with respect to} $Y|\bm{X}$ following the distribution $P(Y|\bm{X};\bm{\theta}^{(0)})$. This is generally valid because of Shannon's Lemma~\citep{shannon1948mathematical}, which states that if the model is correct and identifiable, the true parameter vector $\bm{\theta}^{(0)}$ \textit{uniquely maximizes} the expected log-likelihood, $E_{\bm{\theta}^{(0)}}[\log{P(Y|\bm{X};\bm{\theta})}]$. For ease of exposition, we develop the approach for the setting in which maximum likelihood estimation~(MLE) is used to fit the model. However, for complex models some form of regularization is virtually always used as a modification of the log-likelihood function. With minor modifications discussed in Section~\ref{ss_CD:high_dim_score}, the approach still applies to regularized MLE.

Given a parametric distribution or {marginal} likelihood function {$P(Y | \bm {X}; \bm{\theta})$} for an individual observation $(\bm{X}, Y)$, and assuming we have an i.i.d. training data set $\{(\bm {x}_i, y_i)\}_{i=1}^n$ drawn from this distribution, the score vector for $ (\bm {x}_i, y_i)$ is defined as 
\begin{align}
\bm{s}(\bm { \theta}; (\bm {x}_i, y_i)) = \nabla _{\bm { \theta}} { \log{P(y_i | \bm {x}_i; \bm{\theta})}}
\label{eqn_CD:score_func}
\end{align}
where $\nabla _{\bm { \theta}}$ is the derivative operator {with respect to} the vector of parameters, $\bm {\theta}$. From fundamental theory~(Proposition 3.4.4 from~\citep{bickel2015mathematical}), under certain regularity conditions, if the model is correct and we have a true parameter vector {$\bm { \theta} ^{ (0)}$}, the expected score vector evaluated at $\bm { \theta} ^{ (0)}$ is zero:
\begin{align}
E_{\bm { \theta} ^{ (0)}}[\bm{s}(\bm { \theta}^{ (0)};(\bm {X}, Y))|\bm {X}] = \int\bm{s}(\bm { \theta}^{ (0)};(\bm {X}, Y=y))P(Y=y|\bm{X};\bm{\theta}^{(0)})dy = \bm{0}.
\label{eqn_CD:score_exp_zero}
\end{align}
In other words, the conditional expectation of the score vector evaluated at $\bm { \theta} ^{ (0)}$ is identically zero.

Concept drift is defined as a change in $ P (Y|\bm {X})$, and in our parametric model setting it equates to a change in the parameters of the conditional distribution $P(Y|\bm {X}; \bm { \theta})$. This and Equation~(\ref{eqn_CD:score_exp_zero}) suggest that a general approach for monitoring for concept drift is to monitor for changes in the mean of the sample score vector~(\ref{eqn_CD:score_func}). In particular, with no concept drift, we have $\bm{\theta}=\bm{\theta}^{(0)}$, so that the sample score vector $\bm{s}(\bm { \theta}^{ (0)};(\bm {X}, Y))$ is zero-mean from the above discussion. In contrast, if there is concept drift, this means $\bm{\theta}$ has changed from $\bm{\theta}^{(0)}$ to some other value $\bm{\theta}^{(1)} \neq \bm{\theta}^{(0)}$, in which case the new mean of the score vector $E_{\bm { \theta} ^{ (1)}}[\bm{s}(\bm { \theta}^{ (0)};(\bm {X}, Y))|\bm {X}] \neq \bm{0}$ is no longer zero~(for small changes in $\bm{\theta}$ and under certain identifiability assumptions). As a preview to how we will apply the above concepts in practice~(see Section~\ref{ss_CD:sgd_score} for details), a set of estimated parameters from a training data set will take the place of $\bm { \theta}^{ (0)}$, and to whatever the current values of $\bm{\theta}$ have drifted will take the place of $\bm{\theta}^{(1)}$.

To illustrate the above arguments more concretely, consider the generalized linear model (GLM)~\citep{nelder1972generalized}, which encompasses many models commonly used in applications. The canonical form of the GLM model and score vector are
\begin{align}
\begin{aligned}
P(Y|\bm{X};\bm { \theta}, \phi) =& \exp\big\{\frac{Y \psi(\bm{X};\bm { \theta})-b( \psi(\bm{X};\bm { \theta}))}{ a ( \phi)} + c(Y; \phi)\big\}\text{,~and} \\
\bm {s}(\bm { \theta};(\bm {X}, Y)) =& \frac{1}{a( \phi)}(Y - b'( \psi (\bm{X};\bm { \theta} )))\nabla _{ \bm { \theta}} \psi(\bm{X};\bm { \theta} ),
\end{aligned}
\label{eqn_CD:score_glm}
\end{align}
for functions $b(\cdot)$, $ \psi(\cdot)$, $a(\cdot)$, and $c(\cdot)$ and dispersion parameter $\phi$. The conditional mean is $E[Y|\bm{X}]=b'(\psi(\bm{X};\bm{\theta}))$, where the derivative is {with respect to} $\psi$, and the canonical link function $g(\cdot)$ is defined such that $g(E[Y|\bm{X}])=\psi(\bm{X};\bm{\theta})$($=\bm{X}^T\bm{\theta}$ for the traditional GLM). Using a Taylor expansion, the expected score vector for $\psi(\bm{X};\bm{\theta})=\bm{X}^T\bm{\theta}$ with a small parameter change $ \Delta \bm { \theta}= \bm { \theta}^{(1)}-\bm { \theta}^{(0)}$ is
\begin{align}
\begin{aligned}
E _{\bm { \theta} ^{(1)}}[\bm {s}(\bm { \theta} ^{(0)};(\bm {X}, Y))|\bm{X}] \approx& \frac{1}{a ( \phi)}b''( \psi(\bm{X}; \bm { \theta} ^{(0)}))\nabla _{ \bm { \theta}} \psi (\bm{X}; \bm { \theta} ^{ (0)}) \nabla _{\bm { \theta}}^T \psi (\bm{X}; \bm { \theta} ^{ (0)}) \Delta \bm { \theta} \\
=& \frac{1}{a ( \phi)}b''( \psi(\bm{X}; \bm { \theta} ^{(0)}))\bm {X}\bm {X}^T \Delta \bm { \theta}.
\end{aligned}
\label{eqn_CD:exp_score_glm}
\end{align}
It is known that $Var[Y|\bm{X}] = a ( \phi)b''(\psi (\bm{X}; \bm { \theta} ))$, so that $b''(\psi (\bm{X}; \bm { \theta} ))>0$ always holds if $Y$ is not a deterministic function of $\bm{X}$. If we further take the expectation of Equation~(\ref{eqn_CD:exp_score_glm}) {with respect to} the distribution of $\bm{X}$, then the resulting matrix that premultiplies $\Delta\bm{\theta}$ in~(\ref{eqn_CD:exp_score_glm}) is always positive definite if the distribution of $\bm{X}$ is not degenerate. In other words, whenever concept drift happens~(i.e., $\Delta\bm{\theta}\neq \bm{0}$), the mean of the score vector is non-zero. 

Notice that this approximation~(\ref{eqn_CD:exp_score_glm}) does not require $ \psi (\bm{X}; \bm { \theta})$ to be a linear function of $\bm { \theta}$, if certain identifiability conditions on $\bm{\theta}$ are satisfied. For example, $ \psi (\bm{X}; \bm { \theta})$ can be a neural network, in which case $\bm { \theta}$ is the vector of weights and bias parameters for the neural net. In later studies with real and simulated data sets in Section~\ref{s_CD:real_data} and {supplementary Section~C}, we empirically show that our score-based approach is also effective for neural networks.

In summary, for GLM-type response models with linear or nonlinear $\psi(\cdot)$ that satisfy certain identifiability conditions on $\bm{\theta}$, \textit{the mean of the score vector changes if and only if $P(Y|\bm{X};\bm{\theta})$ changes}, at least for sufficiently small changes in $\bm{\theta}$ (for large changes in $\bm{\theta}$ in most practical settings, the mean of the score vector is likely to change substantially, although it is not clear if any mathematical theory exists to guarantee this). This is an important property that provides the theoretical basis for our score-based concept drift monitoring and diagnosis. Note that error-based methods do not enjoy this property, i.e., $P(Y|\bm{X};\bm{\theta})$ can change in ways that do not change the error rate~{(see supplementary Section~B.1 for a detailed discussion and a concrete example)}.

\subsection{Interpretations with Empirical Data and Incorrect Models}
\label{ss_CD:sgd_score}
The zero-mean property $E_{\bm{\theta}^{(0)}}[\log{P(Y|\bm{X}; \bm{\theta}^{(0)})}] = \textbf{0}$ of the score vector and the uniqueness of the parameters $\bm{\theta}$ that maximize the expected log-likelihood $E_{\bm{\theta}^{(0)}}[\log P(Y|\bm{X}; \bm{\theta})]$ are guaranteed to hold only when the model is correct; that is, when the supervised learning model $P(Y|\bm{X};\bm{\theta})$ is of the same structure as the true predictive relationship $P(Y|\bm{X})$. Recalling the adage that ``All models are wrong, but some are useful"~\citep{box1976science}, one might wonder to what extent the results in the previous section are applicable when the structure of the model $P(Y|\bm{X};\bm{\theta})$ differs from the true $P(Y|\bm{X})$. A related question is what should we take to be the empirical counterpart to $E_{\bm{\theta}}[\bm{s}(\bm{\theta}^{(0)};(\bm{X},Y) )]$ when $\bm{\theta}^{(0)}$ is replaced by its estimate from a sample of training data, and the expectation is replaced by a sample average over a set of new data or over the same training data. We address both of these issues in this section and also relate the empirical counterpart to SGD for computational reasons. 

Regardless of whether the model structure is correct, in analogy with Equation~(\ref{eqn_CD:score_exp_zero}) we always have
\begin{align}
\begin{aligned}
&\hat{E}_{(0)}[\bm{s}(\hat{\bm{\theta}}^{(0)};(\bm{X}, Y))] \vcentcolon=\frac{1}{n}\sum_{i=1}^{n}\bm{s}(\hat{\bm{\theta}}^{(0)};(\bm{x}_i, y_i))=\bm{0}, \text{where} \\
&\hat{\bm{\theta}}^{(0)} \vcentcolon =  \argmax_{\bm{\theta}}\hat{E}_{(0)}[\log{P(Y|\bm{X}; \bm{\theta})}] \vcentcolon= \argmax_{\bm{\theta}}\frac{1}{n}\sum_{i=1}^n \log{P(y_i|\bm{x}_i;\bm{\theta})},
\end{aligned}
\label{eqn_CD:score_exp_zero_emp}
\end{align}   
the operator $\hat{E}_{(0)}$ denotes a sample average over the training data $\{(\bm{x}_i, y_i)\}_{i=1}^n$, and $\hat{\bm{\theta}}^{(0)}$ is the MLE of $\bm{\theta}^{(0)}$ for the training data. That is, when we fit a model using MLE, the gradient of the training log-likelihood is identically zero, i.e., $\nabla_{\bm{\theta}}\hat{E}_{(0)}[\log{P(Y|\bm{X}; {\bm{\theta}})}]|_{\bm{\theta}=\hat{\bm{\theta}}^{(0)}} \vcentcolon=\nabla_{\bm{\theta}}\frac{1}{n}\sum_{i=1}^n \log{P(y_i|\bm{x}_i;{\bm{\theta}})|_{\bm{\theta}=\hat{\bm{\theta}}^{(0)}}}=\frac{1}{n}\sum_{i=1}^n\bm{s}(\hat{\bm{\theta}}^{(0)};(\bm{x}_i, y_i)) = \hat{E}_{(0)}[\bm{s}(\hat{\bm{\theta}}^{(0)};(\bm{X}, Y))]=\bm{0}$, even if the model is not the correct structure. Thus, (\ref{eqn_CD:score_exp_zero_emp}) is the empirical counterpart of (\ref{eqn_CD:score_exp_zero}) with the estimated $\hat{\bm{\theta}}^{(0)}$ taking the place of the true $\bm{\theta}^{(0)}$. 

Now suppose the true predictive relationship $\tilde{P}(\tilde{Y}|\tilde{\bm{X}})$ changes from $P(Y|\bm{X})$ over some new set of data $\{(\tilde{\bm{x}}_i, \tilde{y}_i)\}_{i=1}^{\tilde{n}}$. In this case a different set of parameters $\hat{\bm{\theta}}^{(1)} \neq \hat{\bm{\theta}}^{(0)}$ for the same supervised learning model structure $P(Y|\bm{X};\bm{\theta})$ will generally provide a better fit to the new data than did $\hat{\bm{\theta}}^{(0)}$, where 
\begin{align}
\begin{aligned}
\hat{\bm{\theta}}^{(1)} \vcentcolon= \argmax_{\bm{\theta}}\hat{E}_{(1)}[\log{P(\tilde{Y}|\tilde{\bm{X}}; \bm{\theta})}] \vcentcolon= \argmax_{\bm{\theta}}\frac{1}{\tilde{n}}\sum_{i=1}^{\tilde{n}} \log{P(\tilde{y}_i|\tilde{\bm{x}}_i;\bm{\theta})},
\end{aligned}
\label{eqn_CD:score_exp_nonzero_emp}
\end{align}   
and the operator $\hat{E}_{(1)}$ denotes the sample average over the new data. Thus, the gradient $\nabla_{\bm{\theta}}\hat{E}_{(1)}[\log{P(\tilde{Y} | \tilde{\bm{X}}; {\bm{\theta}})}]|_{\bm{\theta}=\hat{\bm{\theta}}^{(0)}} \vcentcolon= \nabla_{\bm{\theta}}\frac{1}{\tilde{n}}\sum_{i=1}^{\tilde{n}} \log{P(\tilde{y}_i | \tilde{\bm{x}}_i; {\bm{\theta}})}|_{\bm{\theta}=\hat{\bm{\theta}}^{(0)}} = \frac{1}{\tilde{n}} \sum_{i=1}^{\tilde{n}} \bm{s}(\hat{\bm{\theta}}^{(0)};(\tilde{\bm{x}}_i, \tilde{y}_i)) = \hat{E}_{(1)}[\bm{s}(\hat{\bm{\theta}}^{(0)};(\tilde{\bm{X}}, \tilde{Y}))]$ of the log-likelihood for the new data~(but with the gradient evaluated at the original estimate $\hat{\bm{\theta}}^{(0)}$) will generally differ from zero. The more $\tilde{P}(\tilde{Y}|\tilde{\bm{X}})$ changes from $P(Y|\bm{X})$, the more we expect $\hat{\bm{\theta}}^{(1)}$ to differ from $\hat{\bm{\theta}}^{(0)}$, and the more we expect the new average score vector $\hat{E}_{(1)}[\bm{s}(\hat{\bm{\theta}}^{(0)};(\tilde{\bm{X}}, \tilde{Y}))]$ to differ from $\bm{0}$. This provides the justification for our score-based concept drift monitoring approach, which tracks the mean of the score vector $\bm{s}(\hat{\bm{\theta}}^{(0)};(\bm{X}_i, Y_i)) = \nabla_{\bm{\theta}}\log P(Y_i|\bm{X}_i; \bm{\theta})|_{\bm{\theta}=\hat{\bm{\theta}}^{(0)}}$ to detect and analyze changes in it. In supplementary Section~B.3, we provide an example investigating the effect of fitting an incorrect model structure on the performance of our score-based approach and demonstrate that it still performs substantially better than the residual-based EWMA in this case.

If the supervised learning model $P(Y|\bm{X};\bm{\theta})$ is of the same structure as the true predictive relationship $P(Y|\bm{X})$, both $P(Y|\bm{X})$ and $P(\bm{X})$ are constant across the training data $\{(\bm{x}_i, y_i)\}_{i=1}^n$, and $n \to \infty$, then under some regularity conditions the MLE $\hat{\bm { \theta}} ^{ (0)}$ is consistent and  $\hat{E}_{(0)} [\bm{s}(\hat{\bm { \theta}} ^{ (0)};(\bm {X}, Y))] \to E_{\bm { \theta} ^{ (0)}}[\bm{s}(\hat{\bm { \theta}} ^{ (0)};(\bm {X}, Y))] \to E_{\bm { \theta} ^{ (0)}}[\bm{s}(\bm { \theta} ^{ (0)};(\bm {X}, Y))] = \bm {0}$. In this case, there is no distinction between the theoretical version of the score-based monitoring arguments and their empirical version discussed above. With large $n$, SGD is often used to fit models to the training data, which involves approximating the gradient of the log-likelihood function using individual training observations or mini-batches of training observations at each iteration of the optimization algorithm. The SGD estimator of $\bm { \theta} ^{ (0)}$ converges to the batch MLE under certain conditions involving step size and other considerations~(see, e.g., Theorem 4.7 of \citep{bottou2018optimization}). In this case, under the same asymptotic conditions stated above, the SGD estimator $\hat {\bm { \theta}}_{\mathrm{SGD}}$ is also consistent and $\hat{E}_{(0)} [\bm{s}(\hat{\bm { \theta}}_{\mathrm{SGD}};(\bm {X}, Y))] \to \bm{0}$ as $n \to \infty$. 

The score-based approach does not add much extra cost to the current framework of training and using models. For retrospective analysis, the sample score vectors are a byproduct of the SGD~(or related optimizers, e.g., ADAM~\citep{kingma2014adam}), since the mini-batch gradients are of the form $\nabla _{\bm { \theta}} \sum _{i=1} ^{m} \log{P(y_i|\bm {x}_i;\bm{\theta})} = \sum _{i=1} ^{m} \bm{s}(\bm { \theta};(\bm {x}_i, y_i))$, where $m$ is the batch size. Even for prospective analysis, prediction of new data usually partially calculates the score vectors. For example, prediction for neural networks requires forward-propagation, and another backward-propagation in memory would generate the score vector. Thus, we do not need much extra computation to apply the score-based approach.

With finite training data size $n$ and finite new sample sizes $\tilde{n}$ for monitoring~(including $\tilde{n}=1$), noise in $\bm{s}(\bm { \theta};(\bm {x}_i, y_i))$ and its sample averages must be considered. In particular, we need to distinguish by how much $\bm{s}(\hat{\bm { \theta}}^{(0)};(\bm {x}_i, y_i))$~(or its sample average over some moving time window) should differ from $\bm{0}$ before we conclude that $P(Y|\bm{X})$ has changed. The MEWMA monitoring strategy in the next section is designed to distinguish this type of noise from legitimate changes in $P(Y|\bm{X})$. Moreover, for models fitted with regularization, the gradient of the log-likelihood itself is not zero over the training data, because the regularization penalty is included in the optimization objective function. Regardless, the score-based monitoring method can still be applied with the minor modification to the score vectors discussed in Section~\ref{ss_CD:high_dim_score}.

\subsection{An MEWMA Approach for Monitoring the Score Vector}
\label{ss_CD:MEWMA}
As discussed in the previous sections, monitoring for concept drift reduces to monitoring for changes over time in the mean of the score vector. Among other challenges, this requires distinguishing between noise in the score vectors for individual observations vs. an actual mean change. The MEWMA~\citep{lowry1992multivariate} has emerged as one of the most effective techniques for detecting changes in the mean vectors in general, and in this section we develop it for monitoring the mean of the score vector.

The MEWMA at time $t$, denoted by $\bm{z}_t$, is defined recursively~(for $t=1,2,\cdots$) via:
\begin{align}
\bm {z}_t = \lambda \bm {s}_t + (1 - \lambda) \bm {z} _{t-1},
\label{eqn_CD:ewma}
\end{align}
where $\bm {s}_t$ is the to-be-monitored random vector at time $t$, which in our case is the score vector $\bm {s}_t \vcentcolon= \bm{s}(\hat{\bm { \theta}}^{(0)};(\bm {x}_t, y_t))$; $ \lambda$ is a weighting parameter; and $\bm{z}_0$ can be initialized as the sample mean of $\bm {s}_t$ over some small initial batch of observations. An equivalent expression for the recursive relationship~(\ref{eqn_CD:ewma}) is $\bm {z}_t = \lambda\sum _{\tau=1}^t (1-\lambda) ^{t-\tau} \bm{s} _{\tau} + (1-\lambda)^t \bm{z}_0$, which gives exponentially decaying weights to the historic observations up to $\bm{s}_t$.

Since $\bm{s}_t$ and $\bm{z}_t$ are vectors, and we desire to detect changes in the mean of $\bm{s}_t$ in any direction, our MEWMA approach monitors the Hotelling $T^2$~\citep{hotelling1947multivariate} statistic 
\begin{align}
T_t^2 = (\bm {z}_t-\bar { \bm {s}})^T \widehat {\bm { \Sigma}} ^{-1}(\bm {z}_t-\bar { \bm {s}}),
\label{eqn_CD:hotellingt2}
\end{align}
where $\widehat {\bm { \Sigma}}$ and $\bar {\bm{s}}$ are the covariance matrix and sample mean vector of $\bm {s}_t$, respectively, estimated from the training and Phase-I data, as described in later parts of this section. In the statistical process control (SPC) literature, Phase-I refers to a retrospective analysis of previously collected data, with the goals of statistically characterizing the process behavior and establishing control limits to be used in Phase-II. Phase-II refers to a prospective analysis in which one monitors (usually in real-time) new observations as they arrive with the goal of detecting any change in the process behavior as quickly as possible after it occurs. If $T_t^2$ at some $t$ exceeds a specified upper control limit~($UCL$) determined as described later, this is taken to be an indication that $P(Y|\bm{X})$ in the time-vicinity of observation $t$ has changed from what it was when $\hat{\bm{\theta}}^{(0)}$ was estimated. The convention that we recommend and that we have used in all of our later examples in the paper is to estimate $\bar {\bm{s}}$ from only the Phase-I data and $\widehat {\bm { \Sigma}}$ from only the training data. The reason we do not use any Phase-I data to estimate $\widehat {\bm { \Sigma}}$ is that the control limits are determined from the Phase-I data, and estimating $\widehat {\bm { \Sigma}}$ from the same Phase-I data can result in a form of overfitting and a $UCL$ that is too small, which results in too many false alarms. The reason we do not use any training data to estimate $\bar {\bm{s}}$ is that if there is concept drift between the training and Phase-I data and we use the training data to estimate $\bar { \bm {s}}$, then $\bm {z}_t-\bar { \bm {s}}$ will not be zero-mean over the Phase-I data, which can result in a $UCL$ that is too large, which reduces the sensitivity to concept drift in Phase-II. Note that if there is concept drift between the training and Phase-I data, this can be detected via our retrospective analysis, as illustrated in the later examples. 

The score-based concept drift monitoring framework is divided into three steps, which will be illustrated in more detail in the later examples. In Step 1, we collect a batch of data in time order, $\{\bm {x}_i, y_i\} _{i=1} ^{m}$, where $m$ denotes the sample size of this batch. Then, after fitting a preliminary supervised learning model to these data, a retrospective analysis is conducted by applying the MEWMA to these same data. If the MEWMA indicates these data are stable and no significant concept drift is detected, we proceed to Step $2$. On the other hand, if the MEWMA detects significant concept drift in these data then one can discard a portion of the data prior to the concept drift in an attempt to ensure that the remaining data are stable. For other kinds of concept drift, like gradual drifting or seasonality as in the bike-sharing example in {supplementary Section~B.4}, we can try to modify the model by incorporating other covariates to reduce the concept drift. This retrospective analysis would then be repeated~(perhaps multiple times) on the remaining data to verify whether it was stable vs. experienced concept drift. When the remaining data are deemed stable, then we proceed to Step $2$.

Let $n\leq m$ denote the size of the stable data from Step $1$ and denote these data by $\mathcal{D} \vcentcolon=\{\bm {x}_i, y_i\} _{i=1} ^{n}$. The purpose of Step $2$ is to establish the $UCL$  and then to prospectively monitor new data for concept drift, as the new data are collected. To establish the $UCL$, we divide $\mathcal{D}$ into two parts: $ \mathcal{D}_1 \vcentcolon= \{\bm {x}_i, y_i\} _{i=1} ^{n_1}$ and $\mathcal{D}_2 \vcentcolon= \{\bm {x}_i, y_i\} _{i=n_1+1} ^{n}$. The first part, $\mathcal{D}_1$, is used to retrain the parametric supervised learning model of interest, and the second part, $\mathcal{D}_2$, is used to establish the $UCL$ in Phase-I. In the Phase-I analysis, the score vectors for the $\mathcal{D}_2$ data are computed based on the model fitted to the $\mathcal{D}_1$ data. The MEWMA statistics $\{\bm{z}_t\}_{t=n_1+1}^n$ and $T^2$ statistics $\{T_t^2\}_{t=n_1+1}^n$ are then computed for the $\mathcal{D}_2$ data. The user specifies a desired false alarm probability $\alpha$~(e.g., $\alpha=0.0001$, $\alpha=0.001$, etc.), and the $UCL$ is set as the $1-\alpha$ sample quantile of $\{T_t^2\}_{t=n_1+1}^n$. After computing the $UCL$ in Phase-I, in Phase-II the MEWMA with this $UCL$ is applied prospectively to the sample score vectors for a new set of ``on-line" data $\tilde{\mathcal{D}}\vcentcolon=\{\tilde{\bm{x}}_i,\tilde{y}_i\}_{i = 1}^{\tilde{n}}$~(i.e., new data for which the fitted supervised learning model is to be used to predict the response, e.g., new credit card applicants who are being scored for credit risk by the fitted model). The purpose of the Phase-II analysis is to detect as quickly as possible if concept drift occurs in the new data, so that the supervised learning model can be updated accordingly. 

The purpose of Step $3$ is to conduct a diagnostic analysis to help determine the nature of the concept drift, if any drift is detected in Phase-II. This involves plotting the $T_t^2$ statistic vs. $t$ over the data set of interest. Note that the $T^2$ statistic aggregates changes in any of the components of the score vector into a single scalar statistic. To provide richer diagnostic information and help understand which parameters have changed, analogous univariate EWMA control charts for~(transformed) individual components of the score vector should also be constructed. We describe these univariate control charts in Section~\ref{s_CD:decou_cd}.

The Step $3$ diagnostic procedure can also be used in a purely retrospective analysis following Step $1$, if it is desired to understand the nature of the nonstationarity in $P(Y|\bm{X};\bm{\theta})$ over any set of training data to which a supervised learning model is fitted.

For the other methods based on scalar metrics~(e.g., error rates) to which we compare our approach and for the transformed components of the score vector that we describe in Section~\ref{s_CD:decou_cd}, we use univariate EWMA~\citep{roberts1959control} control charts because the EWMA in Equation~(\ref{eqn_CD:ewma}) is a scalar in this case. One main difference between a univariate EWMA and an MEWMA is that the MEWMA $T^2$ statistic aggregates the changes in the mean vector into a single scalar statistic, and only larger values of $T^2$ indicate a change in the mean from $\bar{\bm{s}}$. Thus only an $UCL$ is needed in the MEWMA. In contrast, the univariate version of the EWMA in Equation~(\ref{eqn_CD:ewma}) to detect changes in the mean of a scalar random variable are two-sided in nature and have a lower control limit~($LCL$) as well as a $UCL$. Changes in the mean are indicated by the univariate EWMA statistic falling either below the $LCL$ or above the $UCL$. After concept drift is detected, users can also attempt to validate whether the alarm was truly due to concept drift or to some other phenomenon such as covariate drift using methods discussed in~\citep{raza2015ewma} and \citep{mejri2018new, mejri2021new}, for example.

\subsection{Handling High-Dimensional and Regularized Models}
\label{ss_CD:high_dim_score}
Many machine learning models are becoming increasingly complex, and some state-of-the-art models can have millions of parameters, e.g., deep neural networks. With such high-dimensional parameters, the sample covariance matrix $\widehat {\bm { \Sigma}}$ in Equation~(\ref{eqn_CD:hotellingt2}), which must be inverted, is very likely to be singular or close to it. For example, when the sample size of our training data~(denoted by $n_1$ as in Section~\ref{ss_CD:MEWMA}) is smaller than the number of parameters, the sample covariance is always singular. 

%The preceding solution would be model specific in that the chosen subset of parameters would depend on the structure of the model. 
{With very high-dimensional $\bm{\theta}$ like in convolutional neural networks (CNNs), one simple solution is to monitor the elements of the score vector corresponding to only a specific subset of the parameters, which we further discuss in the next section. An alternative} solution is to modify the covariance matrix to circumvent the problem of inverting a singular or poorly-conditioned matrix. One way to accomplish this is to add a nugget parameter~(borrowing terminology from Gaussian process modeling) to all diagonal entries of $\widehat {\bm { \Sigma}}$. Specifically, for some $\delta>0$, we substitute $\widetilde {\bm { \Sigma}} \vcentcolon= \widehat {\bm { \Sigma}}+ \delta \mathbf {I}$ for $\widehat {\bm { \Sigma}}$ in Equation~(\ref{eqn_CD:hotellingt2}). To understand the effect of this nugget parameter, denote the eigen-decomposition of the sample covariance matrix as $\widehat {\bm { \Sigma}} = \mathbf {Q}\bm { \Lambda} \mathbf {Q}^T$, where $\mathbf {Q}$ is an orthogonal matrix of eigenvectors and $\bm{\Lambda}=diag\{\lambda_1, \lambda_2,\cdots, \lambda_{q}\}$, (where $q=dim(\bm{\theta})$), is a diagonal matrix of eigenvalues, and suppose the eigenvalues are arranged in non-increasing order. Then, we can write the approximated sample covariance matrix as $\widetilde {\bm { \Sigma}} = \mathbf {Q}\widetilde{\bm { \Lambda}} \mathbf {Q}^T$, where $\widetilde{\bm { \Lambda}} \vcentcolon= \bm { \Lambda} + \delta \mathbf {I}$. Using $\widetilde {\bm { \Sigma}}$ in place of $\widehat {\bm { \Sigma}}$ in Equation~(\ref{eqn_CD:hotellingt2}) suppresses unimportant directions of variation. The resulting MEWMA with this modified covariance matrix would not have the issue of ill-conditioning. 

Another way to accomplish this is to use a pseudo-inverse of $\widehat {\bm { \Sigma}}$ in Equation~(\ref{eqn_CD:hotellingt2}) instead of its actual inverse, after dropping any eigenvalues $\lambda_i$ with $\lambda_1/\lambda_i>\gamma$ for some specified maximum condition number $\gamma$. Define the diagonal matrix $\bm { \Lambda} ^{-}=diag\{\lambda_1^{-1},\lambda_2^{-1},\cdots,\lambda_k^{-1},0,\\ \cdots,0\}$ where $\lambda_k$ is the smallest eigenvalue with $\lambda_1/\lambda_k$ no greater than $\gamma$. Then, the pseudo-inverse is defined as $\widehat {\bm { \Sigma}} ^{-} = \mathbf {Q}\bm { \Lambda}^{-}\mathbf {Q}^T$. This is equivalent to applying principal component analysis to $\bm{s}_t$ and retraining only the principal component directions in which the variation in $\bm{s}_t$ is not negligible. In our approach we used the nugget parameter instead of the pseudo-inverse, so as to enable detection in changes in the mean of the score vector in all directions. For all of our examples, we chose the nugget parameter as the smallest value of $\delta$ for which the condition number of $\widehat{\bm{\Sigma}}+\delta\mathbf{I}$ was no larger than $10^4$.

Another related issue is regularization of complex models, which is almost always required to combat overfitting. A regularization term $J(\bm{\theta})$ is used to penalize complex models and large parameters by fitting the model to minimize the regularized loss function $l(\bm{\theta}) = -\sum_{i=1}^n \log P(y_i|\bm{x}_i;\bm{\theta}) + J(\bm{\theta}) = - \sum_{i=1}^n \big\{ \log P(y_i|\bm{x}_i;\bm{\theta}) - \frac{J(\bm{\theta})}{n}\big\}$, instead of the MLE loss function $-\sum_{i=1}^n \log P(y_i|\bm{x}_i;\bm{\theta})$. For example, for $L_2$ regularization, we use $J(\bm{\theta}) = \frac{c}{2}||\bm{\theta}||_2^2$, where $c>0$ is the regularization parameter. The gradient of the regularized loss function becomes $\nabla_{\bm{\theta}}l(\bm{\theta}) = - \sum_{i=1}^n \big\{ \bm{s}(\bm{\theta};(\bm{x}_i,y_i))-\frac{\nabla_{\bm{\theta}}J(\bm{\theta})}{n} \big\}$. Suppose we redefine the score vector as $\bm{s}(\bm{\theta};(\bm{x}_i,y_i)) \leftarrow \bm{s}(\bm{\theta};(\bm{x}_i,y_i)) -\frac{\nabla_{\bm{\theta}}J(\bm{\theta})}{n}$ and view this as the regularized score vector. Using the same arguments as in the unregularized situation, it follows that the regularized score vector is zero-mean when there is no concept drift, and so it still makes sense to monitor for changes in the mean of the regularized score vector. The MEWMA and Hotelling $T^2$ computations in the concept drift monitoring procedure are exactly the same but with the score vector replaced by the regularized score vector. For $L_2$ regularization, this amounts to replacing $\bm{s}(\bm{\theta};(\bm{x}_i,y_i)) \leftarrow \bm{s}(\bm{\theta};(\bm{x}_i,y_i)) - \frac{c}{n}\bm{\theta}$. Since the score vector and the regularized score vector only differ by the constant vector $\frac{\nabla_{\bm{\theta}}J(\bm{\theta})}{n}$, the monitoring statistic remains unchanged when regularization is used, because the centered MEWMA $\bm {z}_t-\bar { \bm {s}}$ and the covariance matrix $\widehat {\bm{\Sigma}}$ in Equation~(\ref{eqn_CD:hotellingt2}) are translation-invariant. Another way to view the above is that we are replacing the likelihood with its Bayesian counterpart $P(Y|\bm{X};\bm{\theta})\exp\big\{-\frac{J(\bm{\theta})}{n}\big\}$, which incorporates a prior distribution~(a Gaussian prior for $L_2$ regularization) on the parameters.

Note that for $L_1$ regularization, which is commonly used for linear and logistic regression modeling (i.e., LASSO regression), $J(\bm{\theta})$ is not differentiable with respect to the parameters that have been regularized to zero. Similar to what was described above for the case of high-dimensional $\bm{\theta}$, one could take the score vector for only the subset of parameters that are non-zero. In this case, it is not clear whether changes in the parameters that were regularized to zero would be reflected in a change in the mean of the partial score vector. This warrants future study. {In the online supplementary Section~D, we discuss additional extensions of the approach to supervised learning models with extremely high-dimensional parameters (e.g., deep learning networks) and to non-differentiable models (e.g., tree-based models).}

\section{Diagnostics and Enhanced Monitoring of Individual Components}
\label{s_CD:decou_cd}
The Hotelling $T^2$ statistic aggregates mean shifts in the components of the score vector into a single scalar statistic. In order to provide diagnostic insight into the nature of the change in $P(Y| \bm {X}, \bm{\theta})$~(e.g., which parameters have changed) and/or to enhance the ability of the procedure to detect some changes, it is helpful to monitor individual components of the score vector $\bm{s}(\bm { \theta}^{ (0)}; (\bm {X}, Y))$ or the transformed version described below. To construct a univariate EWMA chart for the $j^{\mathrm{th}}$ component ($j=1,2,\cdots,q$) of the score vector, denoted by $s_{j,t} \vcentcolon= [\bm{s}(\hat{\bm { \theta}}^{(0)};(\bm {x}_t, y_t))]_j$, the univariate counterpart of Equation (\ref{eqn_CD:ewma}) is

\begin{align}
z_{j,t} = \lambda s_{j,t} + (1 - \lambda) z_{j,t-1}.
\label{eqn_CD:uniewma}
\end{align}
In this case, $z_{j,t}$ is plotted directly on the univariate EWMA chart with both a $LCL$ and $UCL$. The chart signals a change in the mean of the $j^{\mathrm{th}}$ component at observation $t$ if $z_{j,t}$ falls either below the $LCL$ or above the $UCL$. If $\lambda$ is small, $z_{j,t}$ is approximately normal by the central limit theorem, and one can set  $\{LCL_j,UCL_j\} = \bar{z}_j \pm z_{\alpha/2}SD[z_j]$, where $SD[z_j]$ and $\bar{z}_j$ denote the standard deviation and sample average over the training data, $\{z_{j,t}\}_{t=1}^{n_1}$, and Phase-I data, of $\{z_{j,t}\}_{t=n_1+1}^n$, $z_{\alpha/2}$ is the upper $\alpha/2$ quantile of the standard normal distribution, and $\alpha$ is the desired false alarm rate. Alternatively, if the Phase-I sample size $n-n_1$ is sufficiently large, one can choose $LCL_j$ and $UCL_j$ directly as the lower and upper $\alpha/2$ quantiles of the empirical distribution of $\{z_{j,t}\}_{t=n_1+1}^n$. 

If the goal is to diagnose and isolate which parameter(s) have changed, then it is preferable to replace the score components in the univariate EWMA by the components of a transformed version of the score vector discussed below, to effectively decouple the changes in the individual parameters. To illustrate this coupling, consider a linear Gaussian regression model $Y = \bm{X}^T\bm{\theta} + \epsilon$ with $\epsilon$ following a zero-mean Gaussian distribution with variance $\sigma^2$, the score vector for which is $\bm{s}(\bm { \theta}; (\bm {X}, Y)) = \frac{(Y - \bm{X}^T\bm{\theta})\bm{X}}{\sigma^2}$. With no concept drift~(i.e., $\bm { \theta} = \bm { \theta}^{ (0)}$), the mean of the score vector is $E_{\bm{ \theta}^{ (0)}}[\bm{s}(\bm { \theta}^{ (0)}; (\bm {X}, Y))] = E[E_{\bm{ \theta}^{ (0)}}[ \frac{(Y - \bm{X}^T\bm{\theta}^{ (0)})\bm{X}}{\sigma^2}|\bm {X}]] =  E[\bm{0}] = \bm{0}$. After concept drift, suppose the parameters change to $\bm { \theta} ^{ (1)}$, and denote this change by $ \Delta \bm { \theta} = \bm { \theta} ^{ (1)} - \bm { \theta}^ { (0)}$. The mean of the score vector after the concept drift is $E_{\bm{ \theta}^{ (1)}}[\bm{s}(\bm { \theta}^{ (0)}; (\bm {X}, Y))]=E[E_{\bm{ \theta}^{ (1)}}[\frac{(Y - \bm {X}^T\bm { \theta}^{ (1)} + \bm {X}^T\Delta \bm { \theta}) \bm {X}}{\sigma^2} |\bm {X}]] = E[E_{\bm{ \theta}^{ (1)}}[\frac{\bm {X}\bm {X}^T\Delta \bm { \theta}}{\sigma^2} |\bm {X}]] = \frac{E [\bm {X}\bm {X}^T] \Delta \bm { \theta}}{\sigma^2}$. Thus, we can decouple the changes in the parameters by premultiplying the score vector mean by the inverse of the expected Fisher information matrix $\frac{E [\bm {X}\bm {X}^T]}{\sigma^2}$, i.e., via the transformation $ \Delta \bm { \theta} = \sigma^2 \big(E [\bm {X}\bm {X}^T]\big)^{-1} E_{\bm{ \theta}^{ (1)}}[\bm{s}(\bm { \theta}^{ (0)}; (\bm {X}, Y))]$. 

{
%In general, we have two forms of implementation for decoupling the parameter change $\Delta \bm { \theta}$~(see details in supplementary Section~A): 
For more general models and $P(Y | \bm{X}; \bm{\theta})$ we consider two different forms of decoupling, depending on whether one is interested in decoupling the components of $\bm{\theta}$ or (for regularized models) the components of the regularized version $\tilde{\bm{\theta}}$ of the true parameters, where $\tilde{\bm{\theta}}$ denotes the minimizer of the regularized version of the population loss function $E_{\bm{\theta}} [-log P(Y | \bm{X}; \bm{\theta}) + J(\bm{\theta})/n]$. The appropriate decoupling equations are  (see supplementary Section~A for derivations):
\begin{align}
\begin{aligned}
 \Delta \bm { \theta} = \mathbb {I}^{-1}(\bm { \theta}^{ (0)}) E_{\bm{ \theta}^{ (1)}}[\bm{s}(\tilde{ \bm { \theta}} ^{ (0)}; (\bm {X}, Y))],
\end{aligned}
\label{eqn_CD:decouple}
\end{align}
and
\begin{align}
\begin{aligned}
  \Delta \tilde{\bm { \theta}} = \widetilde{\mathbb {I}}^{-1}(\bm { \theta}^{ (0)}) E_{\bm{ \theta}^{ (1)}}[\bm{s}(\tilde{ \bm { \theta}} ^{ (0)}; (\bm {X}, Y))], 
\end{aligned}
\label{eqn_CD:decouple-2}
\end{align}
where $\mathbb {I}(\bm { \theta}^{ (0)}) \vcentcolon= E_{\bm { \theta}^{(0)}} \big[ \bm{s}( \tilde{ \bm { \theta}} ^{ (0)}; (\bm {X}, Y)) \bm{s}^T(\tilde{ \bm { \theta}} ^{ (0)};(\bm{X},Y)) \big]$ is the expected Fisher information matrix (which can be estimated as the sample covariance matrix of $\bm{s}(\tilde{ \bm { \theta}} ^{ (0)}; (\bm {X}, Y))$ over the training data prior to concept drift), $\widetilde{\mathbb {I}}(\bm { \theta}^{ (0)}) \vcentcolon= - E_{\bm { \theta}^{ (0)}} \big[ \nabla_{\bm{\theta}} \nabla^T_{\bm{\theta}} \big\{ \log{P(Y | X; \bm{\theta})} - \frac{J(\bm{\theta})}{n} \big\} |_{\bm{\theta} = \tilde{\bm{\theta}}^{(0)}} \big]$ is an alternative expression for the expected Fisher information matrix with regularization which can be estimated as the sample average of the Hessian matrix $-\nabla_{\bm{\theta}} \nabla^T_{\bm{\theta}} \big\{ \log{P(Y | X; \bm{\theta})} - \frac{J(\bm{\theta})}{n} \big\} |_{\bm{\theta} = \tilde{\bm{\theta}}^{(0)}}$ over the training data prior to concept drift. Here, $E_{\bm{ \theta}^{ (1)}} [\bm{s}(\tilde{ \bm { \theta}} ^{ (0)}; (\bm {X}, Y))]$ can be estimated as the sample mean of $\bm{s}(\tilde{ \bm { \theta}} ^{ (0)}; (\bm {X}, Y))$ over some relevant window of data following the concept drift. Alternatively, expressions for the population expectations of the Hessian matrices are available for some models like traditional GLMs~(e.g., linear and logistic regression), in which case these expressions can be used instead of their sample versions. Notice that the two decoupling expressions~(\ref{eqn_CD:decouple}) and~(\ref{eqn_CD:decouple-2}) are different in general. However, if there is no regularization~(i.e., if $J(\bm{\theta}) = 0$ as discussed in Section~\ref{ss_CD:high_dim_score}), then $\tilde{ \bm { \theta}} ^{ (0)} = \bm { \theta} ^{ (0)}$, and it is well known that $\mathbb{I}$ and $\widetilde{\mathbb{I}}$ are equivalent expressions for the expected Fisher information matrix, in which case,~(\ref{eqn_CD:decouple}) and~(\ref{eqn_CD:decouple-2}) are the same. We will refer to either the transformation~(\ref{eqn_CD:decouple}) or~(\ref{eqn_CD:decouple-2}) as ``\textit{Fisher decoupling}".
}

In light of the above Fisher decoupling results, as an alternative to the univariate EWMAs~(\ref{eqn_CD:uniewma}) on the score components, we can construct univariate EWMAs on the decoupled score components
\begin{align}
z_{j,t} = \lambda \tilde{s}_{j,t} + (1 - \lambda) z_{j,t-1},
\label{eqn_CD:uniewma-decoupled}
\end{align}
where $\tilde{s}_{j,t}$ is the $j^{\mathrm{th}}$ component of the decoupled score vector $\tilde{\bm{s}}_t \vcentcolon= \widehat{\mathbb {I}}^{-1}(\bm { \theta}^{ (0)})\bm{s}_t$, where $\widehat{\mathbb{I}}(\bm { \theta}^{ (0)})$ is an estimate of either ${\mathbb {I}}(\bm { \theta}^{ (0)})$ or $\widetilde{\mathbb {I}}(\bm { \theta}^{ (0)})$. The $LCL$ and $UCL$ for~(\ref{eqn_CD:uniewma-decoupled}) are determined analogously to those for~(\ref{eqn_CD:uniewma}), based on the empirical distribution of $\{z_{j,t}\}_{t=n_1+1}^n$.

In addition to providing diagnostic information on which parameter(s) have changed, and how they have changed, the decoupled univariate EWMAs~(\ref{eqn_CD:uniewma-decoupled}) have the following, additional benefit over just using the MEWMA. As discussed earlier, under fairly general conditions, the mean of $\bm{s}(\bm { \theta}^{ (0)}; (\bm {X}, Y))$ changes if and only if $P(Y|\bm{X};\bm{\theta})$ changes. Thus, a change in $P(\bm{X})$ alone with no change in $P(Y|\bm{X};\bm{\theta})$ will not cause the mean of $\bm{s}(\bm { \theta}^{ (0)}; (\bm {X}, Y))$, or the mean of $\bm{z}_t$ in~(\ref{eqn_CD:ewma}), to change. However, a change in $P(\bm{X})$ alone can cause the mean of the Hotelling $T^2$ statistic in~(\ref{eqn_CD:hotellingt2}) to increase~(by changing the covariance of $\bm{s}(\bm { \theta}^{ (0)}; (\bm {X}, Y))$), thus potentially leading to more frequent false alarms if the goal is to signal only when $P(Y|\bm{X};\bm{\theta})$ changes. This is easy to see for the case of the linear Gaussian regression model $Y = \bm{X}^T\bm{\theta} + \epsilon$ with score vector $\bm{s}(\bm { \theta}^{ (0)}; (\bm {X}, Y)) = \frac{(Y - \bm{X}^T\bm{\theta}^{ (0)})\bm{X}}{\sigma^2}$. If $P(Y|\bm{X};\bm{\theta})$ does not change~(i.e., $\bm{\theta}$ remains unchanged at $\bm{\theta}^{ (0)}$), then we still have that $\bm{s}(\bm { \theta}^{ (0)}; (\bm {X}, Y)) = \frac{\epsilon \bm{X}}{\sigma^2}$ is zero mean regardless of whether $P(\bm{X})$ changes, but the covariance of $\epsilon \bm{X}$ can obviously change.   

The univariate EWMAs are more robust to false alarms that are caused by a change in $P(\bm{X})$ and a resulting covariance change in $\bm{s}(\bm { \theta}^{ (0)}; (\bm {X}, Y))$, because they chart the individual $z_{j,t}$ components directly and not some quadratic form like the $T^2$ statistic, and the mean of $z_{j,t}$ changes if and only if $P(Y|\bm{X};\bm{\theta})$ changes. A variance change in $z_{j,t}$ typically would not increase the false alarm rate as much as if its mean changes. Moreover, by visual inspection of the univariate EWMA charts, it is easier to determine if an alarm was due to a mean change or to a variance change in $z_{j,t}$ than with the aggregated $T^2$ statistic in the MEWMA chart.  Because a single aggregated monitoring statistic has other advantages, in practice we recommend using both the MEWMA and the decoupled univariate EWMAs, which we illustrate with the examples in the subsequent sections. 

For classification problems (categorical $Y$), the fitted supervised learning model represents $P(Y|\bm{X};\bm{\theta})$ directly, and changes in $\bm{\theta}$ equate to changes in $P(Y|\bm{X};\bm{\theta})$ if the parametric model structure is appropriate. However, for regression problems, the fitted supervised learning model typically (with squared error loss) represents only the conditional mean $E(Y|\bm{X};\bm{\theta})$, and so changes in $\bm{\theta}$ equate to changes in $E(Y|\bm{X};\bm{\theta})$. If (for example) $Var(Y|\bm{X};\bm{\theta})$ changes but $E(Y|\bm{X};\bm{\theta})$ does not, then the mean of the score vector remains zero (since the same $\bm{\theta}$ still provides optimal prediction of the response). However, similar to the situation described above in which $P(\bm{X})$ changes, changes in $Var(Y|\bm{X};\bm{\theta})$ may result in changes in the covariance of the score vector and affect the false alarm rate of the multivariate EWMA. Developing diagnostic methods that can distinguish this is a topic that warrants future study.

\section{{An Illustrative Example:  Credit Risk Modeling}}
\label{s_CD:real_data}
{In order to demonstrate the performance of our score-based concept drift approach and illustrate its usage, we now} present a real data example, {which involves credit risk modeling with data collected over the period $2003$--$2008$ from a major financial company, during which time the subprime mortgage crisis happened}. In supplementary Section~B.1, we provide a simulation example where concept drift results in no change in expected error rate~(and thus, error-based methods cannot detect it), but our score-based approach is able to effectively detect it. {In supplementary Section~B.2 we provide more comprehensive simulation results comparing the drift detection performances of our score-based approach and alternative existing methods. The main conclusion is that our score-based approach achieves better performance across every example that we considered, often substantially so.} In supplementary Section~B.4, we also provide another real data example, the Capital Bikeshare rental data from $2010$--$2020$ during which time the ``sharing economy" steadily expanded, to demonstrate retrospective and prospective analysis for a regression problem. {In supplementary Section~C we use simulation examples to investigate the performance of the Fisher decoupling approach in diagnosing which parameters have changed and how they have changed.

In the credit risk data set, each row corresponds to a unique credit card customer of a major financial company, and we have $196587$ rows of data in total. The covariates for row/customer $i$ include various customer information~($\bm {x}_i$) available at the time the customer applies for the credit account, and the binary response~($y_i$) indicates whether the customer defaults within the first $9$ months after opening the account. For the purposes of plotting various quantities over time, we associate each row $i$ with a time stamp that is taken to be the day on which the response $y_i$ first becomes available. Specifically, the set of all customers associated with a particular day, which we refer to as their ``entry day" into the data set, are those who opened their account within nine months of that day and defaulted on that day~(in which case they are assigned $y_i=1$), together with those who opened their account exactly nine months prior to that day and did not default~(in which case they are assigned $y_i=0$). The data are imbalanced in the sense that there are $3536$ observations with $y_i = 1$, which accounts for around $1.8\%$ of the data. Consequently, when fitting the model to the training data, we assign a weight of $7.5$ to the minority class~(which corresponds to upsampling the minority class by a factor of $7.5$) so that the weighted percentage of the minority class in training data is around $10\%$. For the training, validation, Phase-I, and Phase-II data sets, the classification errors and score vectors in Equation~(\ref{eqn_CD:score_func}) and any quantities related to them are also using the same weighting scheme. That is, for each minority-class observation, its score vector in the MEWMA equation in Equation~(\ref{eqn_CD:ewma}) is multiplied by $7.5$, its Hessian and score vector are multiplied by $7.5$ when computing the average Hessian~(for Fisher decoupling) and average score vector~(for use in the Hotelling $T^2$ statistic~(\ref{eqn_CD:hotellingt2})), etc. 

These data were originally considered in~\citep{im2012time}, who focused on the same $10$ covariates that we consider in this study: $x_1$~(credit risk score from an earlier model used by the company), $x_2$~(credit bureau risk score), $x_3$~(highest credit limit for open revolving credit accounts), $x_4$~(total balance on all open revolving credit accounts), $x_5$~(balance on the highest-utilization open revolving credit account), $x_6$~(credit limit on the highest-utilization open revolving credit account), $x_7$~(number of inquiries in the last $24$ months), $x_8$~(balance on open mortgages), $x_9$~(categorical variable involving status of savings accounts), and $x_{10}$~(number of inquiries in the last $24$ months, excluding the last two weeks). 

We applied concept drift detection to two supervised learning models:  A logistic regression model and a neural network~(with one hidden layer having $50$ activation nodes). The main reason the company fits models of this nature was to score credit card applicants for risk~(via their predicted probability of defaulting) at the time they apply. For the purpose of evaluating the concept drift detection, we trained and validated both models on the data from $2003$-Jan to $2005$-Dec. The data from $2006$-Jan to $2006$-Dec are used as the Phase-I data to calculate control limits. The remainder of the data, from $2007$-Jan to $2008$-Aug are used as the Phase-II data to test the capability of detecting the concept drift due to the subprime mortgage crisis. Recall that the S\&P $500$ declined by more than $50\%$ over a $15$-month period between the end of $2007$-Dec and the end of $2009$-Mar~(from $1478$ to $683$). The prevailing view is that the major root cause of the stock market decline was the subprime crisis, which had been gradually developing prior to that.

We applied our score-based MEWMA with the goal of detecting concept drift as far in advance of the beginning of the crash~($2007$-Dec) as possible. For comparison, we also applied an EWMA on the (weighted) binary classification error, which requires that the predicted probability $P[Y = 1 | X]$ for each observation be converted to a classified label ($Y = 0$ or $Y = 1$). For this, we used the standard approach of classifying $Y = 1$ if the predicted probability exceeds some threshold, and $Y = 0$ otherwise. We chose this classification probability threshold to be $0.1057$ to match the true positive rate and the true negative rate as closely as possible~(alternatively, if costs of false positives and false negatives were known in advance, we could have chosen the threshold to minimize the total misclassification cost). For both methods, we selected $\lambda = 0.001$, so that the effective window is $\frac{1}{\lambda} = 1000$ customers, which corresponds roughly to one week. Figure~\ref{fig_CD:credit_default} shows the results of both concept drift monitoring methods for both models~(logistic regression and neural network) over the Phase-I and Phase-II data. 

As seen in Figure~\ref{fig_CD:credit_default}, for the logistic regression model, the score-based MEWMA consistently signals~(i.e., the $T^2$ statistic consistently falls above the $UCL$ ) beginning around $2007$-Mar, which is about $9$ months prior to the beginning of the economic crash in $2007$-Dec. In this sense, the score-based approach provides advanced warning that something has changed substantially in the predictive relationship between $Y$ and $\bm{X}$, which could have been indicative that serious economic changes were evolving. At the very least, it would have been an indication that the fitted credit risk scoring model was becoming obsolete and that more effective scoring could be achieved by updating the model. In contrast, the EWMA on the error rate does not consistently signal until much later, around $2008$-Jan, which is $10$ months after the score-based MEWMA began to consistently signal and $1$ month after the economic crash began. The results for the neural network model, which  are shown in the right panels of Figure~\ref{fig_CD:credit_default}, are very similar. 

\begin{figure}[!htbp]
\centering
\includegraphics[width = 0.49\linewidth]{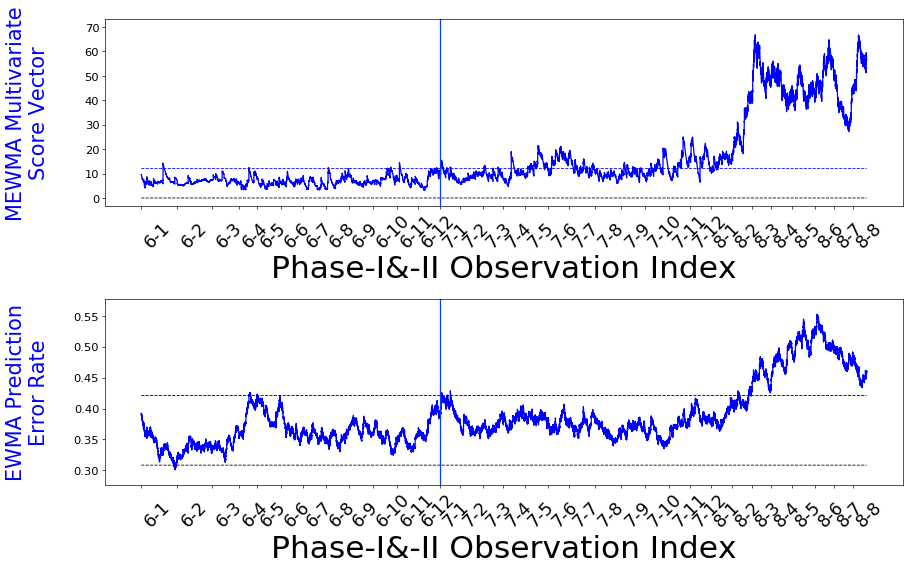}
\includegraphics[width = 0.49\linewidth]{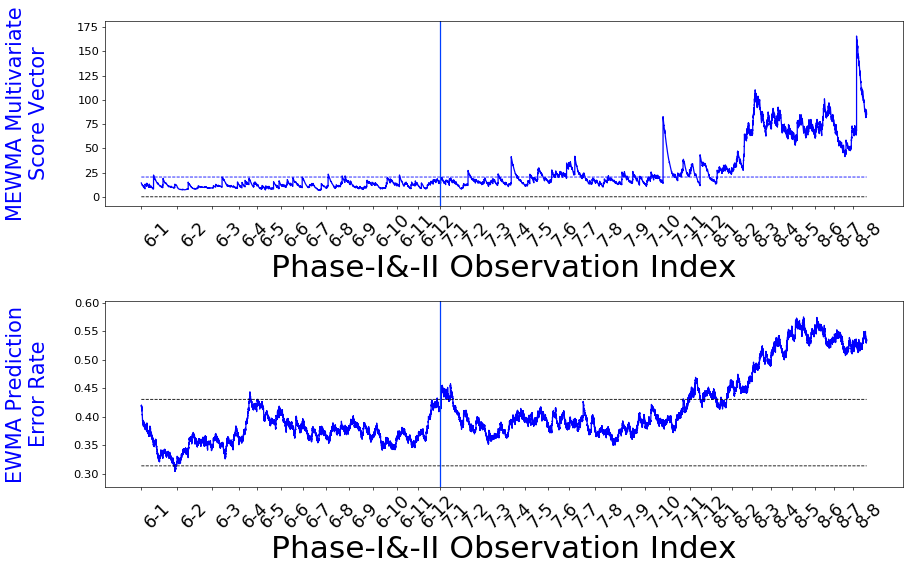}
  \caption{
For the credit risk example, comparison of concept drift monitoring performance for our score-based MEWMA~(the top plots) versus an EWMA on the classification error~(the bottom plots). The left and right plots are for the logistic regression and neural network models, respectively. The horizontal dashed lines are the control limits. The blue vertical line is the boundary between the Phase-I and Phase-II data. The tilted numbers along the horizontal axes below each plot are the year-month indices~(e.g., $6$-$1$ stands for 2006-Jan). The score-based MEWMA consistently signals beginning $2007$-Mar, which is nine months prior to the beginning of the stock market crash in $2007$-Dec. In contrast, the EWMA on the classification error does not consistently signal until around $2008$-Jan. 
}
\label{fig_CD:credit_default}
\end{figure}

Figure~\ref{fig_CD:credit_default_diag} shows several representative univariate component EWMA control charts~(for $\theta_1$, $\theta_2$, $\theta_3$, $\theta_8$, and the intercept $\theta_0$) for the logistic regression model over the Phase-I and Phase-II data. The MEWMA is also shown in the top row as a reference. The component EWMA control charts in the left column are for the original~(coupled) score components~(\ref{eqn_CD:uniewma}), and those in the right column are for the Fisher-decoupled score components~(\ref{eqn_CD:uniewma-decoupled}) using the Hessian matrix version of $\widetilde{\mathbb {I}}(\bm { \theta}^{ (0)})$ for decoupling $\Delta \tilde{\bm { \theta}}$ via~(\ref{eqn_CD:decouple-2}). Rather than estimating the Fisher information matrix as the sample covariance matrix as in Equation~(\ref{eqn_CD:decouple}), we used its theoretical Fisher information matrix expression for logistic regression.

Decoupling the score components reveals different patterns of concept drift than are seen in the original score components. In particular, the decoupled component EWMA for the intercept parameter~(the bottom right plot) shows a clear upwards drift over the entire range of data, whereas this drift is not apparent in the corresponding original component EWMA~(the bottom left plot). The correlation between the intercept term and some of the other covariates has evidently conflated the drift in their corresponding parameters in the left column plots. The upwards drift in the decoupled intercept parameter is telling, as the intercept can be viewed as a regression-adjusted indicator of the overall default rate. Since the drift was upwards, this means that the intercept parameter increased over time, which means that the regression-adjusted likelihood of default~(i.e., the default for applicants having the same covariate values) increased substantially over time. In fact, the concept drift may even be evident earlier in the decoupled component EWMA for $\theta_0$ than in the MEWMA. For example, the decoupled component EWMA for $\theta_0$ falls consistently above the center line~($0$) beginning back in the Phase-I data, around $2006$-Jun. Even though it is not above the $UCL$ at this point, the fact that it is consistently above the center line is an indication that $\theta_0$ has increased. In SPC control charting in general, users typically look for these types of patterns in the charts to signal changes in the mean of what is being charted~\citep{montgomery2007introduction}. 

\begin{figure}[H]
\centering
\includegraphics[width = 0.48\linewidth]{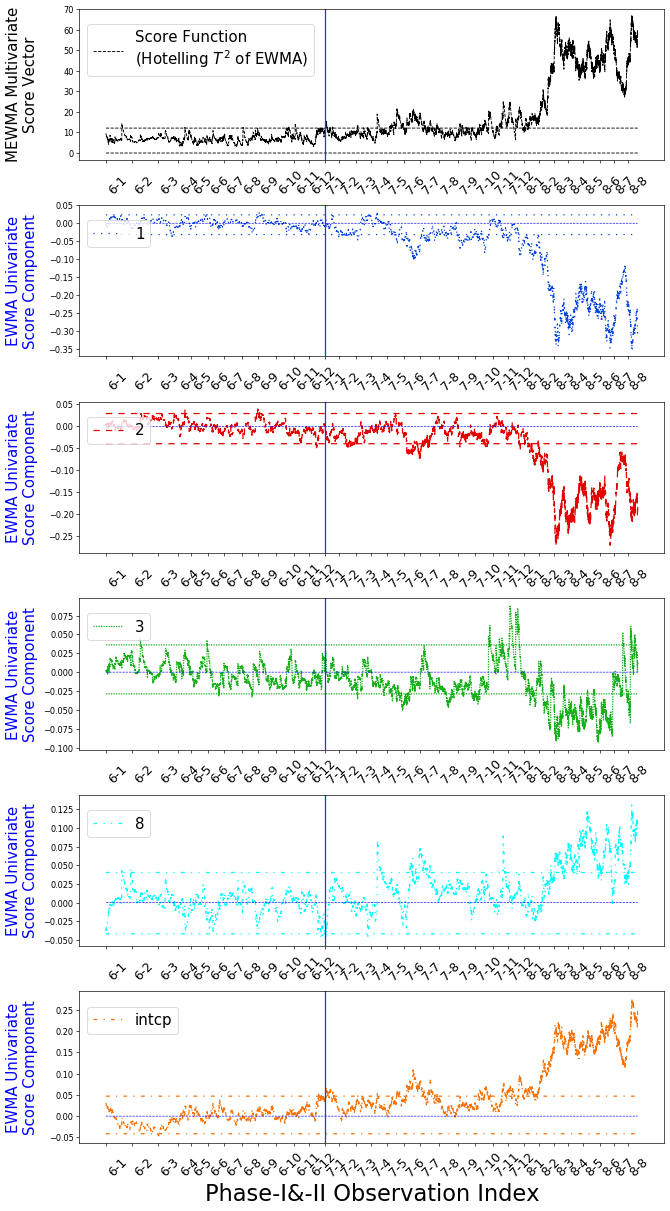}
\includegraphics[width = 0.48\linewidth]{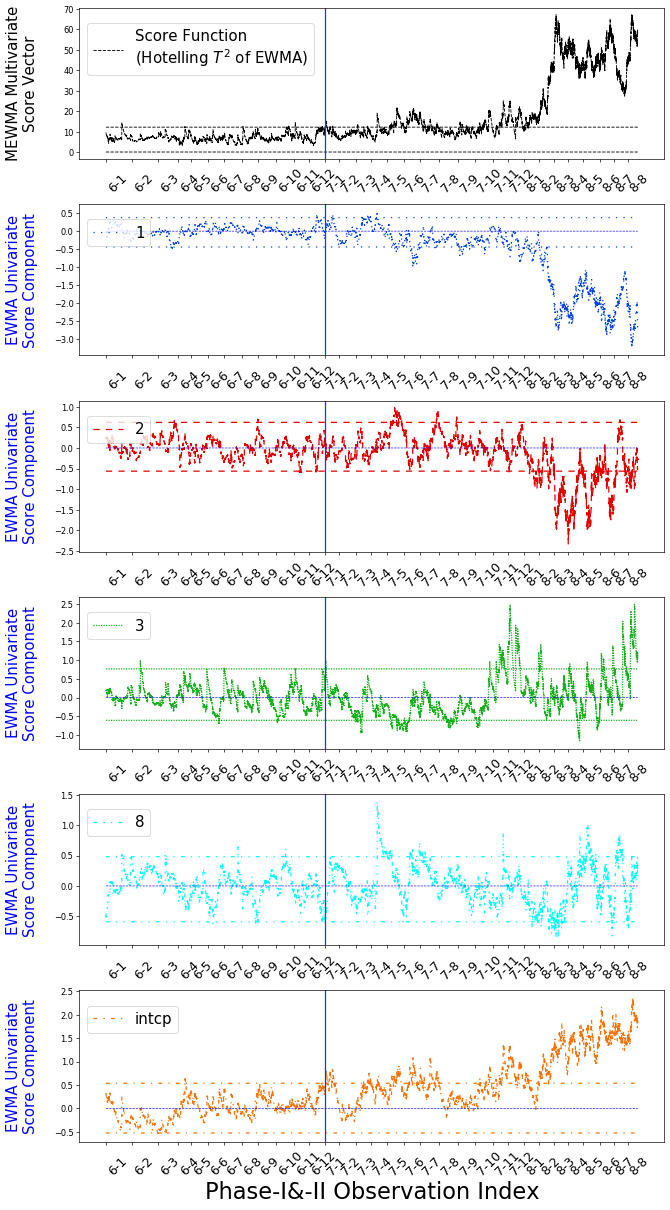}
\caption{
MEWMA~(the top row) and various univariate component EWMA~(the other rows, for $\theta_1$, $\theta_2$, $\theta_3$, $\theta_8$, and the intercept $\theta_0$) control charts for the logistic regression model in the credit risk example over the Phase-I and Phase-II data. The left column are the original score component EWMA control charts~(\ref{eqn_CD:uniewma}), and the right column are the decoupled component EWMA control charts~(\ref{eqn_CD:uniewma-decoupled}). The horizontal lines are the control limits and also a line indicating the zero value for the EWMAs. The blue vertical line is the boundary between the Phase-I and Phase-II data. The decoupled component EWMA control charts show clear drift in a number of parameters, especially the intercept~(the bottom right) and $\theta_1$~(the second from the top, the right column).
}
\label{fig_CD:credit_default_diag}
\end{figure}

The decoupled component EWMA chart for $\theta_1$~(the right column, the $2$nd from top) also shows a clear downwards drift in $\theta_1$, which means that $x_1$~(credit risk score from an earlier model) should be given less weight in predicting credit risk as time evolves. This makes sense, because the earlier model had been fitted to even earlier data.  The decoupled component EWMA chart for $\theta_2$~(the right column, the $3$rd from top) has a different trend that is also less extreme than for $\theta_1$. It is interesting that the component EWMA charts for $\theta_1$ and $\theta_2$ prior to decoupling~(second and third plots in the left column) are far more similar than their decoupled counterparts. This is likely because $x_1$ and $x_2$ have a correlation coefficient of $0.45$, and correlation between covariates conflates the concept drift in their coefficients, whereas the Fisher decoupling is intended to distinguish the concept drift in the coefficients.

\section{Conclusions}
Predictive models are trained on historical data sets, but due to potential changes in the conditional distribution $P (Y|\bm {X})$~(aka, concept drift) the performance of the models may degrade. We have developed a comprehensive, general, and powerful score-based framework for monitoring and diagnosing concept drift. The framework is general in that it applies to {broad classes of} parametric {models}, either regression or classification. It can be used retrospectively~(to analyze whether $P (Y|\bm {X})$ was stable over a past data set and aid in the model building procedure) and prospectively~(to quickly detect changes in $P (Y|\bm {X})$ so that predictive models currently in use can be updated accordingly). We have provided theoretical arguments that, under reasonably general conditions, concept drift occurs if and only if the mean of the score vector changes. Consequently, our score-based procedure is based on monitoring and analyzing changes in the mean of the score vector. For this, we have adopted procedures~(MEWMA and univariate EWMA) that were developed in the SPC literature for monitoring for changes in the mean of multivariate vectors in general. As part of the framework, we have also developed a diagnostic approach that involves decoupling changes in the parameters from the change in the mean of the score vector. 

In simulation and real data examples, we have demonstrated that our score-based monitoring procedure provides much more powerful detection of changes in $P (Y|\bm {X})$ than the state-of-the-art existing approach~(error-based EWMA). In particular, there are examples~(e.g., Figure~B1 in the online supplementary materials) in which the score-based approach quickly detects the change in $P (Y|\bm {X})$, but the error-based approach is completely unable to detect the change because they do not result in a change in the error rate.

We note that although multivariate EWMA monitoring charts (e.g., Figure~\ref{fig_CD:credit_default} and the top row of Figure~\ref{fig_CD:credit_default_diag}) can be used with complex models, the univariate EWMA diagnostic plots (e.g., the remaining rows of Figure~\ref{fig_CD:credit_default_diag}) are intended more for models in which the parameters are meaningful, such as for linear and logistic regression models. However, for complex models like neural networks with many layers, one could construct diagnostic plots for only the parameters in the last layer as described in supplementary Section~D, which represent the effects of features extracted in the previous layers.

One of the assumptions when fitting a supervised learning model of the type considered in this paper is that the training data are stationary over time, in the sense that there is no drift in the predictive relationship between the response variable and the predictor variables. Our concept drift detection approach can be viewed as a way of determining whether this stationarity assumption is violated and, if not, at which point(s) in time the relationship drifted. Moreover, if the univariate diagnostic plots are used, the approach can help indicate which model parameters have drifted.

\section*{Supplementary Materials}
In the online supplementary materials of this article, we provide a file containing derivations of Fisher Decoupling Equations~(\ref{eqn_CD:decouple}) and~(\ref{eqn_CD:decouple-2}), examples demonstrating effectiveness of our methods, simulation study results, and a discussion on potential extensions of our methods. Additionally, the zip file contains python code and data necessary to reproduce some results presented in this paper. 

\section*{Acknowledgements}
This work was funded in part by the Air Force Office of Scientific Research Grant (\#FA9550-18-1-0381), which we gratefully acknowledge. This work was also supported by a startup grant (\#DMR200021) from the Extreme Science and Engineering Discovery Environment (XSEDE)~\citep{towns2014xsede}, which is supported by the National Science Foundation (\#ACI-1548562). We also acknowledge the many helpful comments from the Editor (Dr. Roshan Joseph) and the anonymous Associate Editor and Referees.

\bibliographystyle{Perfect}

\bibliography{sample.bib}
\end{document}